\documentclass[9pt]{extarticle}

\usepackage{spconf,amsmath,graphicx,amssymb}
\usepackage{lipsum}
\usepackage{hyperref}
\usepackage{xcolor}
\usepackage{color}
\usepackage{soul}
\usepackage{bm}
\usepackage{url}
\usepackage{booktabs}
\usepackage{subcaption}
\hypersetup{
    colorlinks=true,
    linkcolor=blue,
    filecolor=magenta,      
    urlcolor=blue,
}

\newcommand\blfootnote[1]{%
  \begingroup
  \renewcommand\thefootnote{}\footnote{#1}%
  \addtocounter{footnote}{-1}%
  \endgroup
}

\title{Progressive Neural Image Compression\\ with Nested Quantization and Latent Ordering}

\name{Yadong Lu$^{*12}$ , Yinhao Zhu$^{*1}$, Yang Yang$^{*1}$, Amir Said$^{1}$, Taco S Cohen$^{3}$}

\address{
$^{1}$Qualcomm AI Research, Qualcomm Technologies, Inc. \\
$^{2}$Department of Statistics,  University of California Irvine \\
$^{3}$Qualcomm AI Research, Qualcomm Technologies Netherlands B.V. \\
}

\begin{document}
\maketitle
\begin{abstract}
% There has been an increasing interest in variable bitrate neural image compression. However, existing learned variable bitrate solutions of image compression need to store separate bitstreams for each quality. 
We present PLONQ, a progressive neural image compression scheme which pushes the boundary of variable bitrate compression by allowing quality scalable coding with a single bitstream. 
% Pronc enables easier rate-control and requires less storage. 
% We present a progressive neural image compression scheme called PLONQ which allows quality scalable coding with a single bitstream.
In contrast to existing learned variable bitrate solutions which produce separate bitstreams for each quality, it enables easier rate-control and requires less storage.
Leveraging the latent scaling based variable bitrate solution, 
we introduce nested quantization, a method that defines multiple quantization levels with nested quantization grids, and progressively refines all latents from the coarsest to the finest quantization level.
To achieve finer progressiveness in between any two quantization levels, latent elements are incrementally refined with an importance ordering defined in the rate-distortion sense.
To the best of our knowledge, PLONQ is the first learning-based progressive image coding scheme and it outperforms SPIHT, a well-known wavelet-based progressive image codec. 

%we introduce nested quantization, a method that progressively refines all latent channels across different quantization levels with nested quantization grid. 
% A single-model variable bit-rate neural image compression scheme is introduced. Progressive coding.
% We present PRONC, a progressive neural image compression scheme via nested quantization and channel ordering. 
%We present a novel variable bit rate image compression algorithm which allows for progressive coding. 
% {\color{red}In contrast to the current learning based image compression approach, our approach is able to achieve two of the most desirable features in image compression simultaneously: 1) training one model and get variable bit rate performance, and 2) successively refine the quality of the reconstructed image through progressive coding. }
% In contrast to existing learned variable bitrate solutions, our approach enables a single high bitrate bitstream to embed all lower bitrate code as its prefix.
% {\color{red}
% At the heart of our approach is a nested quantization method built on top of the existing latent scaling based variable bit rate solution to allow for progressiveness in the coding. 
%We demonstrate through experiments that the rate-distortion performance outperforms BPG at high bitrate regime and is significantly better than existing nested dropout based progressive coding scheme.
% At the heart of our approach is a nested quantization method built on top of existing latent scaling based variable bitrate solution that progressively refines latents across different quantization levels.
% }

\blfootnote{\noindent $^*$ Equal contribution}
\blfootnote{$^1$ Qualcomm AI Research is an initiative of Qualcomm Technologies, Inc.}
\blfootnote{$^2$ Work completed during internship at Qualcomm Technologies Inc.}

\end{abstract}
\begin{keywords}
progressive coding, quality scalable coding, embedded coding, variable bitrate compression, nested quantization
% bit-plane coding, embedded quantization
\end{keywords}

 %\yy{append Nested quantization to the title to make it ProCON?}  \yl{Pronc / Pronco / Proneco ? latent scaling  proseq}
\section{Introduction}
\label{sec:intro}

% \note{1 page for this section (with title and abstract)}
% 1. progressive coding  
% 2. overview for neural image compression + existing variable bitrate solution
% 3. Our contribution  

% \cite{yang2020variablebitrate}
% \cite{choi2019variable}
% \cite{toderici2015variable}
% \cite{johnston2017improved}
% \cite{dosovitskiy2020you}
% \cite{chen2020variable}
% \cite{cui2020gvae}
% \cite{balle2020nonlinear}
% \cite{guo2020variable}
% \cite{zhou2020variable}

Recent developments of learning based lossy image \cite{charm,meanscale,chen2019neural,Cheng_2020_CVPR,BalleLS16a,hyperprior} and video compression \cite{Agustsson_2020_CVPR,ruihan, Lu_2019_CVPR, Habibian_2019_ICCV, Golinski_2020_ACCV} schemes %\cite{BalleLS16a, theis2017lossy, TodericiVJHMSC16, hyperprior} 
%\yy{Yang to update citation}
have witnessed remarkable success in achieving state-of-the-art rate-distortion (R-D) performance compared to traditional codecs in a variety of benchmark datasets. Leveraging the expressiveness of deep auto-encoder networks, these works compress an image by learning a nonlinear transform between the image and the latent, along with a learned entropy model in order to encode the quantized latent. The optimization objective is often framed as a tradeoff between the Rate (R), the amount of bits used to encode the source, and the Distortion (D), the distance between the input source and the reconstructed source: 

\begin{equation}
% \label{rd-formula}
\notag
    \min_{\theta} \mathbb{E}_x \left[ D_{\theta}(x) + \beta R_{\theta}(x) \right]
\end{equation}
where $\beta$ is the tradeoff parameter and $\theta$ represents the model parameters. During training, $\theta$ is typically optimized from end-to-end using stochastic gradient descent. 

\begin{figure}[t]
    \label{flow}
    \centering
    \includegraphics[width=0.45\textwidth]{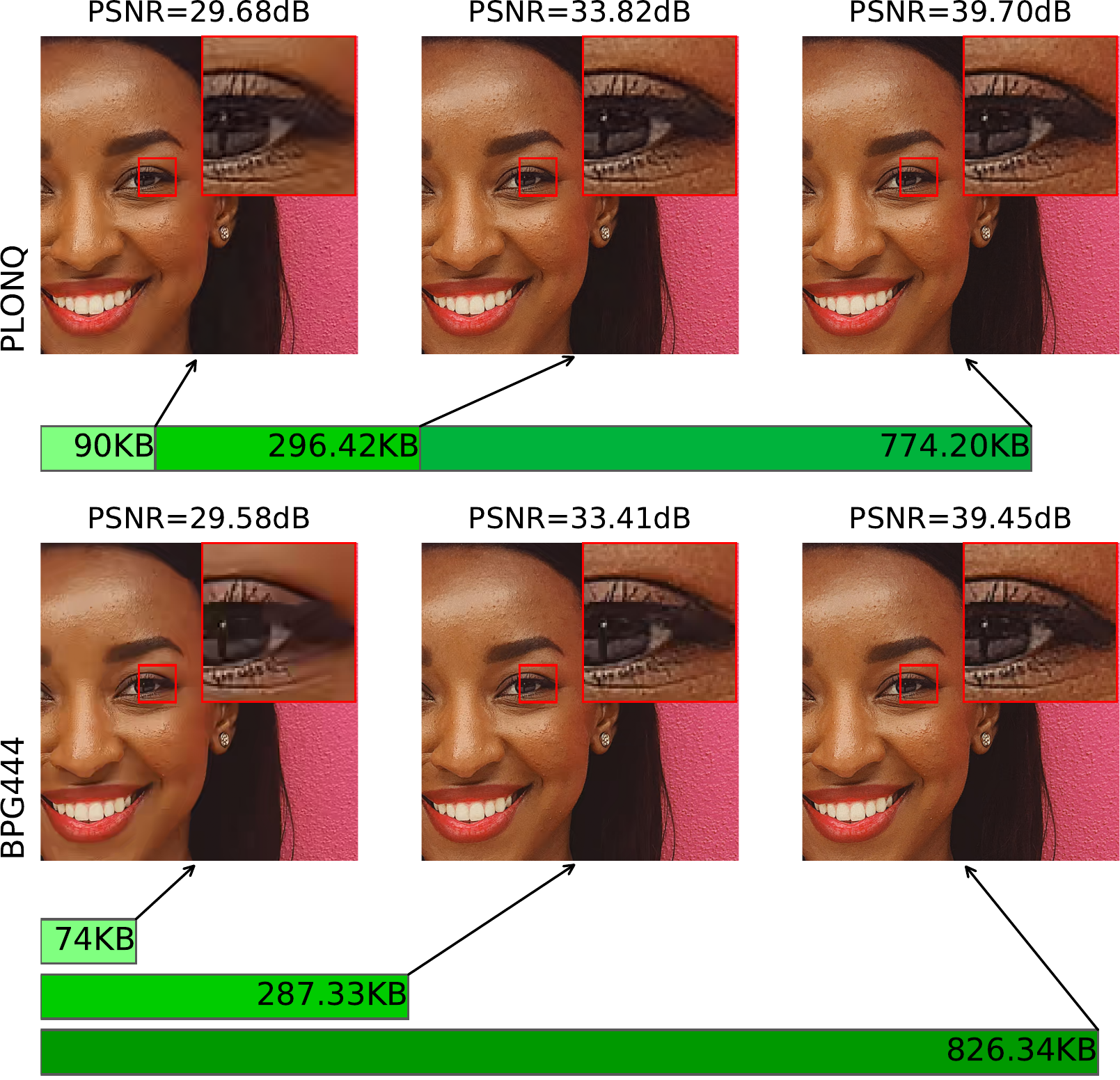}
    \caption{
    Qualitative comparison$^3$ of PLONQ and BPG444. For PLONQ (top), reconstructions at different qualities can be obtained by truncating a single bitstream, whereas for BPG444 (bottom), separate bitstreams have to be stored one for each bitrate option.  
    %\yy{save some white space}
    %\yz{the inset is different on PLONQ and BPG444}
    }
    \vspace{-0.3cm}
\end{figure}
\footnotetext[3]{Displayed image is a cropped version of \texttt{00012\_TE\_1512x2016.png} from JPEG AI testset \cite{jpegai_test}. Reported in this figure is the PNSR and bitrate of the full-size reconstruction. %\yl{Reported in this figure is the PNSR and bitrate of the full-size reconstruction.}

} 

% However, most of these learning based methods are not readily suitable for large scale deployment because: 1) in most of learning based codecs \cite{BalleLS16a, hyperprior}, re-training of model is required to accommodate different bitrate targets, and 2) there is a lack of effective learning based compression scheme allowing for progressive coding. Progressive coding is a highly desirable feature in compression. 
% Progressive coding allows for the easier rate-distortion manipulation. It works for a wide range of bit-rate using a single copy of the latent code encoded in one pass. Therefore it leads to reduced memory by removing the need for storing multiple copies of the latent code in typical variable bitrate solution \cite{cui2020gvae}, which makes it more scalable in deployment.  
% There are different kinds of approaches to progressive coding such as quality progressive and resolution progressive. In our paper, we focus on quality progressive by encoding the image into a successively refined bitstream.  

Most of these learning based methods, however, are not readily suitable for large scale deployment because in most of learning based codecs \cite{charm,meanscale,Cheng_2020_CVPR,BalleLS16a,hyperprior}, separate models need to be trained to accommodate different bitrate targets. To address this issue, Choi et al. \cite{choi2019variable} proposed a conditional auto-encoder to make the parameters in the encoder and decoder network be dependent on the rate distortion trade-off parameter, $\beta$ so that a single model can adapt to different rate-distortion tradeoffs.  More recently, \cite{chen2020variable, cui2020gvae, chen2019neural} proposed latent scaling based variable bitrate compression which is essentially learning to adjust the quantization step size of the latents. After training, the interpolation of the learned scaling parameters can lead to continuous variable bitrate performance. Other attempts \cite{guo2020variable, zhou2020variable} have also been made to address the variable bitrate targets.
% However, none of the above approaches allows for progressive coding.

While variable bitrate solutions make learning based image compression more practical for deployment, they require storing multiple copies of the latent code used for different bitrates, i.e., quality levels \cite{chen2020variable, cui2020gvae}. 
% In contrast, we focus on progressive coding, which enables successive refinement of the reconstruction quality using a single bitstream encoded in a single forward pass, which significantly reduces the storage and simplify the rate control.  
In contrast, with progressive coding, the transmitted bitstream is completely embedded, which be truncated at various points and reconstructed into a series of lower bitrate images \cite{spiht}. Therefore it significantly reduces storage and simplifies rate control.  
% To address this, we develop a progressive coding based compression approach, which works for a wide range of bitrate using a single copy of the latent code encoded in one pass. 
% There are different kinds of approaches to progressive coding such as quality progressive and resolution progressive. In our paper, we focus on quality progressiveness by encoding the image into a bitstream that successively refines the reconstruction quality.
% On top of the existing variable bitrate solution,  Progressive coding is a highly desirable feature in compression. There is a lack of effective learning based compression scheme allowing for progressive coding.  

% Existing literature \cite{toderici2015variable} allows for progressivness by using a encoder based on recurrent neural network 
Existing literature \cite{toderici2015variable} allowing for progressiveness adopted an encoder parameterized by a recurrent neural network, so that it can incrementally transmit the quantized latents. However, the inference time is slow for recurrent neural networks and their performance is not competitive with BPG or the family of hyperprior based solutions \cite{charm, hyperprior}. 
% (cite scale-only, mean-scale, charm). 
In this study, we present \textit{PLONQ, a Progressive coding based neural image compression method via Latent Ordering and Nested Quantization}. PLONQ is built upon the latent scaling based solution while being amenable for deployment. 
% \yy{explain why progressive coding is amenable: (1) encode once, embed-all-bitrate (2) easier rate-control (adjusting rate without re-encoding) (3) reduced storage (4) first step towards scalable neural video compression. These can be moved to the place when we first define progressive coding. } % Most of the recent works using neural network as a for compression focus on getting better compression rate. However, in practice the scalability of the compression methods is one of the main concerns. 
% ordered representation 
% \yy{Do we use scale-only or mean-scale as baseline? @Yadong, can you quickly survey the above papers and see which model is used as baseline for each? For now all the launched experiments are configured to mean-scale.}/
In summary, our work has the following main contributions: 
% \yy{we can quickly mention in the main text here that our work is based off latent-scaling based variable bitrate solution.}
% \vspace{-0.1in}

\begin{enumerate}
% \setlength\itemsep{0em}
% \setlength{\itemsep}{-0.1pt}
    % \item  We provide a comprehensive study of latent scaling based variable bitrate schemes and identify important  problem during optimization is the loss equalization of the different bitrate objectives. \yy{can no longer claim the first point}
    \item We demonstrate the effectiveness of a simple latent scaling based variable bitrate solution using a pretrained high bitrate model without any retraining, dubbed as Naive Scaling.
\vspace{-0.09in}
    %\yy{dubbed should not be followed by as right?}
    % \item {\color{red}We propose a nested quantization method which allows for variable bitrate performance as well as progressiveness in the coding. }
    % \item {\color{blue}We propose nested quantization which produces a embedded bitstream where the earlier codeblocks corresponding to coarser quantization levels.}
    \item We propose nested quantization which enables progressive coding across different quantization levels.
\vspace{-0.09in}
    \item We find sorting latent variables element-wise by prior standard deviation works well to obtain better truncation points between two quantization levels.
    % \yz{not about more or less, it's about better or worse}
%\vspace{-0.09in}
%    \item Combining nested quantization with channel ordering, PLONQ achieves better performance than SPIHT, a well-known progressive coding algorithm, and shows on par performance with BPG at high bitrate.
% \vspace{-0.09in}
    % nested quantization with channel-wise ordering.
\end{enumerate}
% Moved this here as it is a result, not really a contribution (debatable - feel free to move back)
Combining nested quantization with latent ordering, PLONQ achieves better performance than SPIHT, a well-known progressive coding algorithm, and performs on par with BPG at high bitrate.  
\section{Latent Scaling}
\label{sec:scaling}
% \note{0.75 page for this section}

% First we introduce the notations in our approach.  
We base our method on the mean-scale hyperprior model \cite{meanscale} and inherit the notations therein:
%We consider the following rate distortion optimization problem in Eq~\eqref{rd-formula}.
we denote the input image by $x$, and the analysis transform (encoder) and synthesis transform (decoder) by $g_{a}$ and $g_{s}$ respectively.  Both $g_{a}$ and $g_{s}$ are parameterized by convolutional neural networks. The output of the encoder is the continuous latent space tensor $y=g_{a}(x)$, which has a Gaussian prior with its mean $\mu$ and standard derivation $\sigma$ predicted based on hyper latent.
% modeled by a hyper codec. 
%which has a prior distribution of Gaussian with mean $\mu$ and standard derivation $\sigma$,  further modeled by a hyper prior \cite{meanscale}.
% whose prior distribution is modelled by a Gaussian hyperprior \cite{meanscale} that predicts a mean $\mu$ and standard deviation $\sigma$ based on hyper latents.
We use $P(\cdot)$ to denote the probability value on an interval and use $\mathbb{P}(\cdot)$ to denote the discretized probability value.

% And the latent $y$ is typically consisted of multiple feature maps, where we call each feature map as the latent channel.

% Further we denote the latent representation as: $\bm{y} = g_{a}(\bm{x})$, further we have denote the beta to latent scaling mapping function to be $g(\beta, c)$, where $\beta$ is the trade-off parameter in equation \ref{rd-formula}. For simplicity, we will denote the latent scaling to be $g$ hereafter in place of $g(\beta, c)$. 

% \yl{explain smaller rate and distortion} 
The idea of latent scaling~\cite{chen2020variable, cui2020gvae}, is to apply a scaling factor $s$ to the latent $y$. % so that it leads to a different trade-off between the rate to code the latent and the distortion of the reconstructed image. 
Because the quantization step size is fixed, this leads to a different tradeoff between rate and distortion.
% In particular, we apply $\frac{1}{s}$ , where $s>1$, to the latent vector $y$ before latent quantization. 
% it is equivalent to increase the quantization bin width hence lead to a coarser quantization and smaller bit-rate when coding the quantized latent.  
% Several works \cite{chen2020variable, cui2020gvae} achieved variable bit-rate solution through learning a set of $s$ under different rate distortion objectives.  
For ease of exposition we consider a single latent variable, so that both $y$ and $s$ are scalar. The diagram of the latent scaling is depicted in Fig~\ref{flow}. The latent $y$ is scaled with $1/s$, where we restrict $s > 1$. In the quantization step (shown in the blue shaded rectangle in Fig~\ref{flow}, we choose to center the scaled latent $y/s$ by its prior mean $\mu/s$, apply the rounding operator $\lfloor \cdot \rceil$ on $(y-\mu)/s$ (so that the estimated mean $\mu$ learned by the hyper-encoder is on the grid), and then add the offset $\mu/s$ back. The dequantized latent $y(s)$ is obtained by multiplying $s$ after the quantization block, i.e.
\begin{align}
% \label{quant}
\centering
    y(s) \triangleq& \left\lfloor \frac{y - \mu}{s} \right\rceil s + \mu.\notag
\end{align}
The prior probability of $y(s)$ used in entropy coding can be derived from the original prior density by change of variables:% in Eq~\eqref{prior prob}:
% \begin{align}
% \label{prior prob}
%     \mathbb{P}(y(s)) =& \int^{\lfloor (y-\mu)/s \rceil + \mu + 0.5}_{\lfloor (y-\mu)/s \rceil + \mu - 0.5} p_{y/s}(a) da \\
%     \notag =&  \int^{\lfloor (y-\mu)/s \rceil s + \mu + 0.5}_{\lfloor (y-\mu)/s \rceil s  + \mu - 0.5} s p_{y/s}(sa)da \\ 
%     \notag =& \int^{\lfloor (y-\mu)/s \rceil s + \mu  +  0.5s}_{\lfloor (y-\mu)/s \rceil s + \mu - 0.5s} p_{y}(a')da' \\
%     \notag =& CDF_{y}(\lfloor (y-\mu)/s \rceil s + \mu + 0.5s) - \\ \notag & CDF_{y}(\lfloor (y-\mu)/s \rceil s + \mu - 0.5s)
% \end{align} 
\begin{align}
    &\mathbb{P}_{Y(s)}(y(s)) = \mathbb{P}_{\frac{Y(s)}{s}}\left(\frac{y(s)}{s}\right) = \mathbb{P}_{\frac{Y(s)}{s}}\left(\left\lfloor \frac{y-\mu}{s} \right\rceil + \frac{\mu}{s}\right) \notag\\
    % =& P\left( \left\lfloor\frac{y-\mu}{s}\right\rceil+\frac{\mu}{s}+\frac{1}{2} \leq \frac{Y(s)}{s}\leq \left\lfloor\frac{y-\mu}{s}\right\rceil+\frac{\mu}{s}+\frac{1}{2}\right)\notag\\
    % =& P\left( \left\lfloor\frac{y-\mu}{s}\right\rceil+\frac{\mu}{s}+\frac{1}{2} \leq  \frac{y - \mu}{s} + \frac{\mu}{s} \leq \left\lfloor\frac{y-\mu}{s}\right\rceil+\frac{\mu}{s}+\frac{1}{2}\right)\notag\\
    &= \int^{\lfloor (y-\mu)/s \rceil + \frac{1}{2}}_{\lfloor (y-\mu)/s \rceil - \frac{1}{2}} p_{(y-\mu)/s}(u) du 
    % \notag =& \int^{\lfloor (y-\mu)/s \rceil + \frac{\mu}{s} + \frac{1}{2}}_{\lfloor (y-\mu)/s \rceil + \frac{\mu}{s} - \frac{1}{2}} p_{y}(sa + \mu)sda \\
    % \notag =&  \int^{\lfloor (y-\mu)/s \rceil + 0.5}_{\lfloor (y-\mu)/s \rceil - 0.5} s p_{(y-\mu)/s}(sa)da \\ 
    = \int^{y^+(s)}_{y^-(s)} p_{y}(v)dv, \label{prior prob}
    % \notag =& CDF_{y}(\lfloor y/s \rceil s + \mu + 0.5s) - \\ \notag & CDF_{y}(\lfloor y/s \rceil s + \mu - 0.5s)
\end{align} 
% where the upper and lower bound of the effective quantization grid are defined as
% % \begin{align}
% % \label{lower, upper}
% %     y^+(s) \triangleq \left\lfloor \frac{y-\mu}{s} \right\rceil s + \mu + \frac{s}{2}, \hspace{0.1in}
% %     y^-(s) \triangleq \left\lfloor \frac{y-\mu}{s} \right\rceil s +\mu - \frac{s}{2}  
% % \end{align}
% \begin{align}
% % \label{lower, upper}
%     y^+(s) \triangleq y(s) + \frac{s}{2},\hspace{0.1in}
%     y^-(s) \triangleq y(s) - \frac{s}{2} \notag
% \end{align}
where $y^+(s) \triangleq y(s) + \frac{s}{2}$ and $y^-(s) \triangleq y(s) - \frac{s}{2}$ can be interpreted as the upper and lower boundary of the effective quantization bin.

Note in Eq.~\eqref{prior prob}, we first apply the change of variable formula for discrete random variables in the first line, and then apply a change of variable formula for continuous random variables from the second to the third line. According to Eq~\eqref{prior prob}, applying the scaling $s$ is equivalent to changing the quantization bin width from 1 to $s$. Hence increasing the latent scaling value $s$ will lead to larger quantization bin width and hence smaller bitrate in the entropy coding, as well as larger distortion of the reconstructed image. 

% Also note that scaling $s$ is applied to the latent $y$ after it is passed to hyper codec so that the prior parameters $\mu, \sigma$  $y(s)$.

% \yy{rounding is normally denoted as $\lfloor\cdot\rceil$} 

% where $p_{(y-\mu)/s}$ is the density function of $(y-\mu)/s$, $p_{y}$ is the original density function of $y$, and $CDF_{y}$ is the original culmulative distribution function of $y$. In our case, we consider 

\begin{figure}[htb]
    \label{flow}
    \centering
    \includegraphics[width=0.45\textwidth]{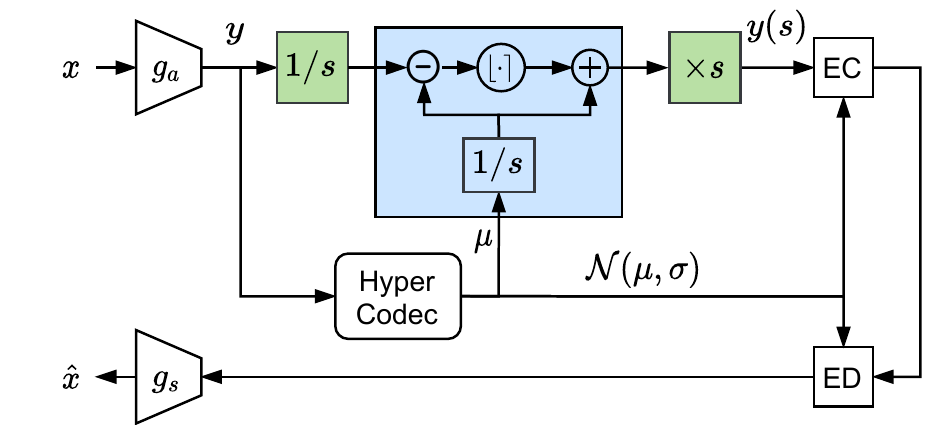}
    \caption{Mean-scale hyperprior with latent scaling. EC is entropy encoder, ED is entropy decoder. Note that scaling $s$ is applied to latent $y$ after being passed to hyper codec so that the prior parameters $\mu, \sigma$ can be shared by $y(s)$ for different $s$.}
    % \yy{Consider shrink the size so that the fontsize look similar to main text}
    % \yy{Can we flatten the figure a bit to save space}
    \label{flow}
\end{figure}

\vspace{-0.25in}

\section{Nested Quantization}
\label{sec:nested}

% \note{0.75 page for this section}

% In the separate scaling approach, we obtained variable bit-rate model through varying the rate-distortion trade-off parameter $\beta$ in Eq~\eqref{rd-formula}, i.e., larger $\beta$ corresponds to larger penalty on rate term, hence smaller number of bits will be used in the coding. During optimization, this is achieved by increasing the scaling magnitude or equivalently increasing the quantization bin width of the latent quantization. Therefore, after training we get an increasing sequence of scaling constants, corresponds to the sequence of rate-distortion parameter $\beta$ of increasing values. In practice, this monotone sequence of scaling constants is closely related to progressiveness in the coding. 

% In this section, we introduce the nested quantization, which is able to leverage the results obtained from the separate scaling to achieve progressiveness in the coding. 
In this section, we introduce nested quantization, one of the key techniques we use to construct a progressive bitstream. 
%As a first step, 
Let us start with a pre-trained mean-scale hyperprior model and
% start with a general example of 
consider
$K$ quantization levels defined by a set of $K$ latent scaling factors $\{s_1, s_2, ..., s_{K}\}$ with $s_1 < s_2 < ... < s_{K}$. We define a scheme, where for each level, the same scaling factor is applied across all latent elements. Effectively, 
%this defines $K$ different bin widths for latent quantization, with which we can obtain $K$ different bitrates with a single pre-trained hyperprior model. 
these $K$ quantization bin sizes translate into $K$ different bitrate options.
We name this simple single-model variable-bitrate scheme as \emph{Naive Scaling}. Despite its simplicity, In Section~\ref{sec:exp} we show that it achieves performance comparable to a more involved scheme~\cite{chen2020variable} that uses $K$ sets of learnable per-channel scaling factors optimized for $K$ Lagrange multipliers $\beta$.

Naturally, the information contained in the discrete latents quantized with different scaling factors follows a total ordering -- the one quantized with smaller bins is always a refinement of the larger.
%and it conveys more information about the true output from the encoder. 
%It is with this observation that we 
%We then
%realize that 
Progressiveness across different quantization levels can then be achieved if we can find a way to embed the information of coarse level quantization into its finer counterpart. Indeed, a principled way to achieve this progressiveness is to encode a finer quantized latent with probabilities conditioned on its coarser quantized values. We term this scheme \emph{nested quantization} and detail it next.

% The idea of nested quantization is to embed the quantization levels into one another so that it allows for progressiveness in the coding, similar to the concept of bit-plane compression.
%\yy{Yang to add more intuitive interpretation behind nested quantization}
% and only two quantization levels as an example showing in Fig~\ref{fig:nested quantization}.  

\begin{figure}[t]
  \centering
  \includegraphics[width=0.85\linewidth]{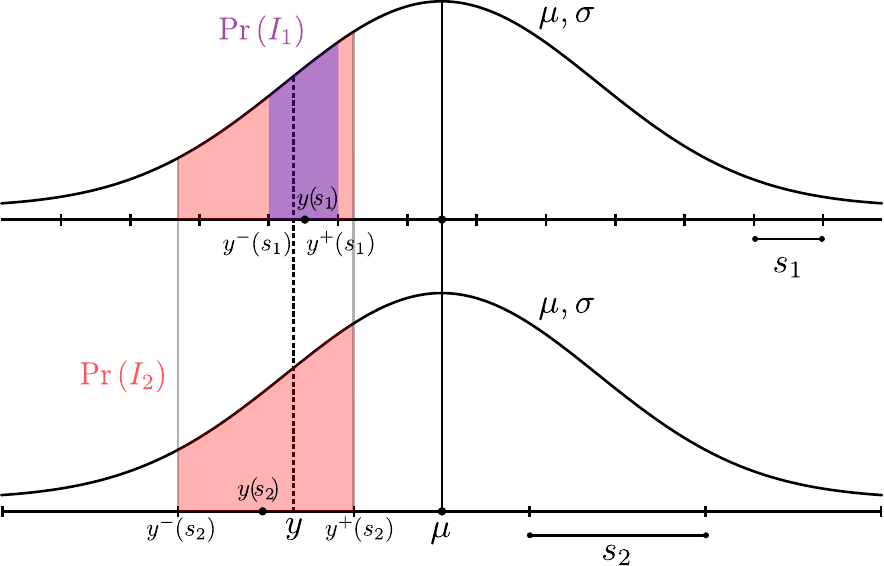}
  \caption{Illustration of nested quantization of $y$ when $K=2$.}
  \label{fig:nested quantization}
\end{figure}

In nested quantization, the encoding happens in $K$ stages. In the first stage, we quantize $y$ with the coarsest bin $s_K$ and encode it with $\mathbb{P}\left(y(s_K)\right)$ as defined in Eq~\eqref{prior prob}. 
%The value of this dicretized probability, $\mathbb{P}\left(y(s_K)\right)$ is equal to the area under the probability density function of the normal distribution $N(\mu, \sigma)$ within the interval $[y^-(s_K), y^+(s_K)]$.
% , where $y^-(s_K), y^+(s_K)$ can be computed using Eq~\eqref{lower, upper}. 
%Note here $\phi$ is the probability density function (pdf) of a standard normal distribution.  Fig~\ref{fig:nested quantization} shows a special case where $K=2$, and the value of $\mathbb{P}\left(y(s_2)\right)$ corresponds to the red shaded area. 
In the remaining stages, we iteratively refine the quantization of $y$ using bin width $s_{k-1}$ that is one level finer than $s_{k}$, for $1\leq k< K$. In particular, we encode an extra piece of information of $y$ quantized to a finer grained level through the conditional probability %$\mathbb{P}\left(y(s_k)|y(s_i)_{i>k}\right)$ 
below, 
\begin{align}
    %\mathbb{P}\left(y(s_K)\right)=&P\left(I_{K} \right),\notag\\
    \mathbb{P}\left(y(s_{k})|y(s_{i})_{i>k}\right)=&P(I_{k})/P(I_{k+1})\text{ for }1\leq k <K,\notag
\end{align}
where $P(I) \triangleq \int_{v\in I} p_y(v)dv$, and $I_k$ is defined below as iterative intersections of quantization bins:
\begin{align*}
    I_{K} &\triangleq \left[y^-(s_K), y^+(s_K)\right], \text{ and }\\
    I_{k} &\triangleq I_{k+1}\cap\left[y^-(s_k), y^+(s_k)\right]\text{ for }1\leq k < K.
    %\\P(I) &\triangleq \int_{v\in I} p_y(v)dv
\end{align*}
% $I_{K} \triangleq \left[y^-(s_K), y^+(s_K)\right]$, $I_{k} \triangleq I_{k+1}\cap\left[y^-(s_k), y^+(s_k)\right]$ for $k < K$, and $P(I) \triangleq \int_{v\in I} p_y(v)dv$.
% Repeat the same procedure in the second step $K-1$ times, we can get a progressively coded latents at $K$ different bit-rates. 
An illustration of this procedure with $K=2$ can be found in Fig~\ref{fig:nested quantization}. The code generated by these $K$ stages form a bitstream that embeds $K$ different bitrates, and the total length of the bitstream is\footnote{Here we assume a perfect entropy coder.} %captured below.
% With a sequence of $N$ scaling constants $[s_1, s_2, ...,  s_{N}]$ obtained from the separate scaling approach, repeating computation of the conditional probability $N-1$ times we get a progressive code at N different bit-rates. 
% \begin{align}
% \label{cond_prob}
%     \mathbb{P}\left(y(s_2)\right)=&P\left(\left[y^-(s_2), y^+(s_2)\right]\right)\\
%     \notag =& \int_{y^-(s_2)}^{y^+(s_2)} \phi(\frac{z-\mu}{\sigma}) dz  \\
%     \mathbb{P}\left(y(s_1)|y(s_2)\right)=&\frac{
%         P\left(\left[y^-(s_2), y^+(s_2)\right]\cap\left[y^-(s_1), y^+(s_1)\right]\right)
%     }{
%         P\left(\left[y^-(s_2), y^+(s_2)\right]\right)
%     }
% \end{align}
% \begin{align*}
%     P(I) = \int_{v\in I} p_y(v)dv
% \end{align*}
% {\color{red} to remove
% \begin{align}
% \label{cond_prob0}
% \centering
%     \mathbb{P}\left(y(s_K)\right)=&P\left(I_{K} \right)= \int_{y^-(s_K)}^{y^+(s_K)} \phi(\frac{a-\mu}{\sigma}) da \\
%     % \notag =& \int_{y^-(s_2)}^{y^+(s_2)} \phi(\frac{z-\mu}{\sigma}) dz  \\
% \label{cond_prob1}
%     \mathbb{P}\left(y(s_{K-1})|y(s_K)\right)=&\frac{
%         P\left(I_{K}\cap\left[y^-(s_{K-1}), y^+(s_{K-1})\right]\right)
%     }{
%         P\left(I_{K}\right)
%     }
% \end{align}}
% In practice, we might get non-uniform discretization intervals due to intersection of the quantization bins. Therefore for every level of quantization, we choose to align its discretization intervals to the intervals 
% defined by the smallest latent scaling value $s_{1}$. In this way, we make sure that the resulting finest quantization level is always identical the level defined by $s_{1}$, therefore it is able to preserve the performance of the highest bit-rate model at the end of the progressive coding procedure. Detailed explanation of the alignment can be found in the Appendix.  
\begin{align*}
    \underbrace{
        \log\frac{1}{\mathbb{P}\left(y\left(s_K\right)\right)}
    }_{
        \substack{
            \text{codeword length w.r.t.}\\
            \text{coarsest quant. level $s_K$}
        }
    }
    + \sum_{1\leq k < K}
    \underbrace{
        \log\frac{1}{
                \mathbb{P}\left(y\left(s_k\right) | y\left(s_{i}\right)_{ i>k}\right)
            }
    }_{
        \substack{
            \text{codeword length w.r.t. refined}\\
            \text{information from } s_{k+1} \text{ to } s_k
        }
    }=
    \log\frac{1}{P(I_1)}.
\end{align*}

There are two issues of this scheme as can be seen from the above equation: (1) conditional probability computation requires tracking of intersected quantization boundaries, which adds to implementation complexity; (2) the sum of codeword length $\log 1/P(I_1)$ is in general larger\footnote{
$P\left(I_1\right)\hspace{-0.1cm}=\hspace{-0.1cm}P\left(\left[y^-\hspace{-0.1cm}\left(s_1\right), y^+\hspace{-0.1cm}\left(s_1\right)\right]\cap I_2\right)
\hspace{-0.1cm}\leq\hspace{-0.1cm}
P\left(\left[y^-\hspace{-0.1cm}\left(s_1\right), y^+\hspace{-0.1cm}\left(s_1\right)\right]\right)
\hspace{-0.1cm}=\hspace{-0.1cm}
\mathbb{P}\left(y\left(s_1\right)\right)$
} than the codeword length with respect to the finest quantization level $\log 1/\mathbb{P}\left(y(s_1)\right)$.

These two issues can be resolved if we consider a set of \emph{fully nested quantization levels}\footnote{This can be viewed as a generalized version of bit-plane coding.} where the set of grid points of the coarser quantization bin is always a subset of the finer. In other words, we can define quantization levels in such a way that $I_{k} = I_{k+1}\cap\left[y^-(s_k), y^+(s_k)\right]= \left[y^-(s_k), y^+(s_k)\right]$, which simplifies the above equation as
\begin{align*}
\log\frac{1}{\mathbb{P}\left(y\left(s_K\right)\right)}
+ \sum_{1\leq k < K}
\log\frac{\mathbb{P}\left(y\left(s_{k+1}\right)\right)}{\mathbb{P}\left(y\left(s_k\right)\right)}
=
\log\frac{1}{\mathbb{P}\left(y\left(s_1\right)\right)}.
\end{align*}
% Note that the final effective codeword length with fully nested quantization levels equals to the codeword length if $y$ were to be directly encoded using the finest grid $s_1$. 
With this we are able to preserve the performance of the highest bit-rate model at the end of the progressive coding procedure.

In practice, we find it beneficial to further generalize this idea to quantization levels with uneven grid, which is applied to PLONQ and explained in Section~\ref{sec:exp}.

% \yy{TODO: add illustration for nested quantization and grid alignment}
% with and without alignment

% \yz{idea like fractional bit-plane coding can be helpful to get more operating points between two beta; what if we use 0.5 as the finest quantization step?}

% \begin{align*}
%     \underbrace{
%         \log\frac{1}{\mathbb{P}\left(y\left(s_K\right)\right)}
%     }_{
%         \substack{
%             \text{codeword length w.r.t.}\\
%             \text{coarest quant. level $s_K$}
%         }
%     }
%     + \sum_{1\leq k < K}
%     \underbrace{
%         \log\frac{\mathbb{P}\left(y\left(s_{k+1}\right)\right)}{\mathbb{P}\left(y\left(s_k\right)\right)}
%     }_{
%         \substack{
%             \text{codeword length w.r.t. refined}\\
%             \text{information from } s_{k+1} \text{ to } s_k
%         }
%     }=
%     \underbrace{
%         \log\frac{1}{\mathbb{P}\left(y\left(s_1\right)\right)}
%     }_{
%         \substack{
%             \text{codeword length w.r.t.}\\
%             \text{finest quant. level $s_1$}
%         }
%     } 
% \end{align*}

% \begin{align*}
%     \log\frac{1}{P\left(I_1\right)} = \log\frac{1}{P\left(I_K\right)} + \sum_{1\leq k < K}\log\frac{P\left(I_{k+1}\right)}{P\left(I_k\right)}
% \end{align*}
\section{Latent Ordering}
\label{sec:ordering}

To obtain more truncation points in the embedded bitstream between two quantization levels, latent variables can be refined  incrementally instead of all at once.
The problem is how to find the optimal order to refine them, so that wherever the resulting bitstream is truncated, the reconstruction has the highest possible quality.
The \textit{embedding principle} states the latent that reduces the distortion the most per bit should be coded first~\cite{taubman2002embedded}. 
% To make it concrete, two key factors for latent ordering are discussed below, namely, coding unit and sorting criterion.

Depending on the embedding granularity, the latent tensor $y$ is divided into coding units $\{y_1, ..., y_N\}$. All the latent elements in one coding unit are refined together at once, corresponding to one truncation point. For the latent tensor with shape $(C, H, W)$\footnote{$C, H, W$ stands for the number of channels (feature maps), height and width of each channel.}, we have considered three types of coding unit for the latent: (1) one channel, (2) one pixel, and (3) one element, which corresponds to a latent slice of size $(1, H, W)$, $(C, 1, 1)$ and $(1, 1, 1)$ respectively. 

Given an order $\rho=(\rho_1, ..., \rho_N)$, ordered coding units $y_\rho=(y_{\rho_1}, ..., y_{\rho_N})$ are refined one by one from scaling $s_k$ to $s_{k-1}$. In mean-scale hyperprior model, the priors of latent elements are independent conditioned on the hyper latent. Thus the bitrate increase of refining $y_{\rho_t}$, i.e. $\Delta R(y_{\rho_t}) = -\log \mathbb{P}\left(y_{\rho_t}(s_{k-1}) | y_{\rho_t}(s_k)\right)$ can be calculated in parallel once the hyper latent is decoded and it is independent of $\rho$. However, with a nonlinear ConvNet decoder, the reduction in distortion depends on the all other ordered coding units, i.e.
% $$\Delta D\left(y_t | y_\rho, s_k, s_{k-1}\right) = D\left(y_\rho(t-1, s_k, s_{k-1} )\right) - D\left(y_\rho(t, s_k, s_{k-1})\right),$$
% where $y_\rho(t, s_k, s_{k-1})=\{y_{\rho[:t]}(s_{k-1}), y_{\rho[t:]}(s_k)\}$, $D(y) $ is the distortion for latent $y$, e.g. MSE$(x, g_s(y))$. 
$\Delta D\left(y_{\rho_t} | y_\rho\right) = D\left(y_\rho(t-1)\right) - D\left(y_\rho(t)\right)$,
where $y_\rho(t)=(y_{\rho_{\leq t}}(s_{k-1}), y_{\rho_{>t}}(s_k))$, and $D(y)=\text{MSE}(x, g_s(y))$ is the distortion for latent $y$.
Thus refining ordered latent by $\rho$ leads to a set of R-D points $\mathcal{H}(\rho) = \{\sum_{i\leq t} \Delta R(y_{\rho_i}), \sum_{i \leq t} \Delta D(y_{\rho_i} | y_\rho)\}_{t=1}^N$. The optimal order $\rho^*$ is the one under which the convex hull of  $\mathcal{H}(\rho^*)$ is  better than that of other orders in the Pareto optimal sense~\cite{taubman2000high}.

% To obtain a finer-grained embedded bitstream between two truncation points achieved by nested quantization, latent variables can be incrementally refined to the finer level instead of at once. The question is how to find the optimal order to code the latent variables, so that wherever the bitstream is truncated, the reconstruction has the highest possible quality. 
% % This is similar to multiple passes in fractional bit-plane coding {\color{red} CITE?}.
% In this section we study two key factors for latent ordering: coding unit and sorting criterion.

% Coding unit is the smallest group of latent elements to be refined together, corresponding to one truncation point in the bitstream. For the latent tensor with shape $(C, H, W)$\footnote{$C, H, W$ stands for the number of channels (feature maps), height and width of each channel.}, we have considered three types of coding unit for the latent, i.e. (1) one channel, (2) one pixel and (3) one element, which corresponds to a latent slice of size $(1, H, W)$, $(C, 1, 1)$ and $(1, 1, 1)$ respectively. 

%  In wavelet-based coding schemes, e.g. EBCOT~\cite{taubman2000high}, the reduced distortion $\Delta D$ and the bitrate $\Delta R$ by coding one unit can be estimated independently and then the ordering problem can be solved by a post-compression R-D optimization.
% , thanks to the linear and orthogonal property of wavelet transforms. 
% However there is no clear relation between the distortion in latent space and image space for neural codec because of its ConvNets-based nonlinear transform. 
Following embedding principle, $\frac{\Delta D}{\Delta R}$ should be sorted in descending order.
We use a greedy procedure to calculate $\frac{\Delta D}{\Delta R}$ under an initial order, e.g. the index order, as the first sorting criterion.
The second criterion is the bitrate $\Delta R$ of a coding unit, because we observe sorting latent channels by $\Delta R$ and by $\frac{\Delta D}{\Delta R}$ lead to similar R-D curve. Since the standard derivation $\sigma$ of the latent prior is correlated to the expected bitrate, it is considered as the third criterion. Note that the first two criteria incur bitrate overhead for coding the order, while the last one does not since $\sigma$ is already known when decoding $y$.
\begin{figure}[htb]
\centering
  \includegraphics[width=0.7\linewidth]{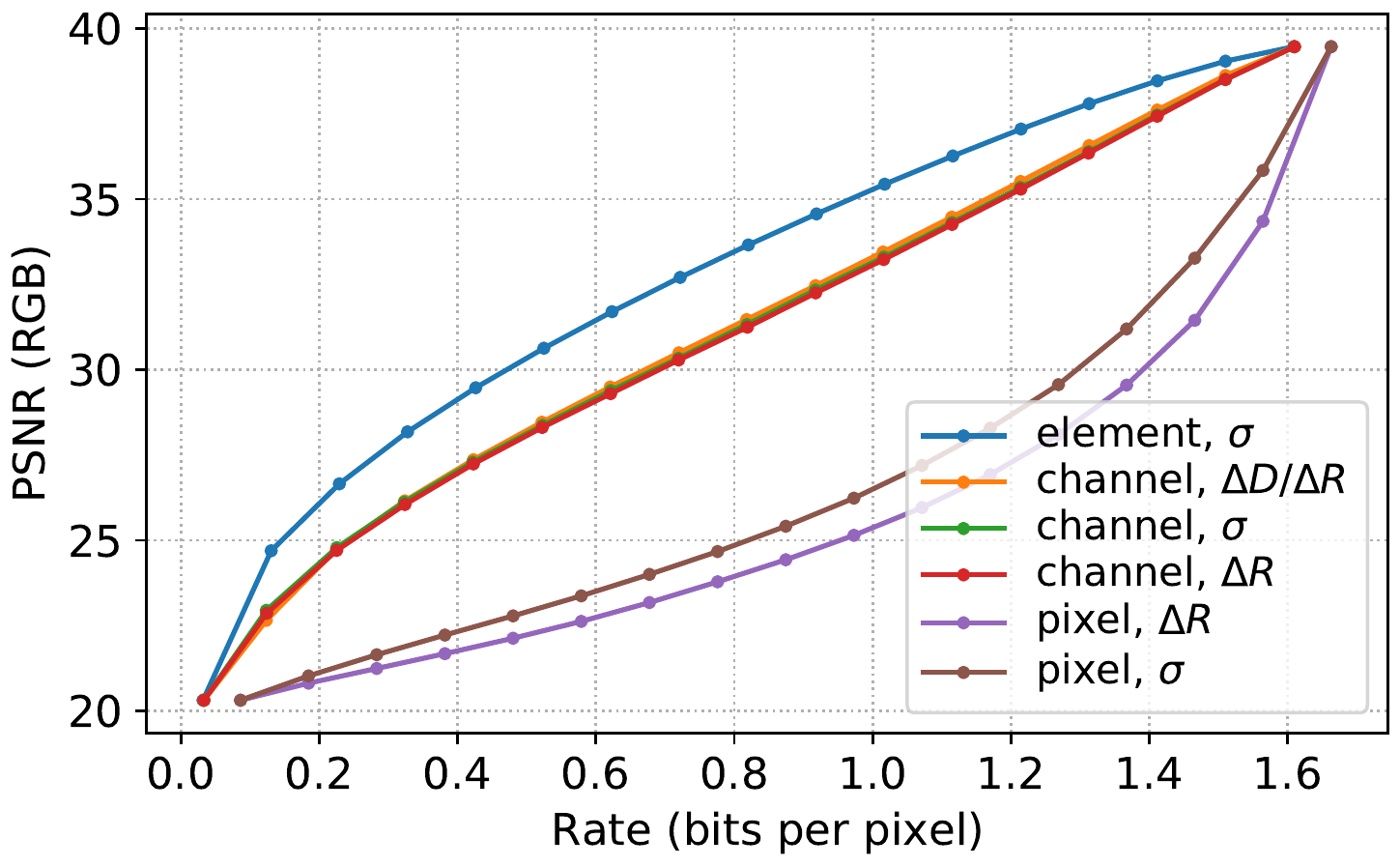}
  \caption{Coding units are incrementally refined between two quantization levels ($s_2=10,000$, i.e. $y(s_2)=\mu$, $s_1=1$) according to a latent order determined by a sorting criterion.}
  \label{fig:latent_ordering}
\end{figure}

\begin{figure*}[t]
\begin{subfigure}{.38\textwidth}
  \centering
  \includegraphics[width=\linewidth]{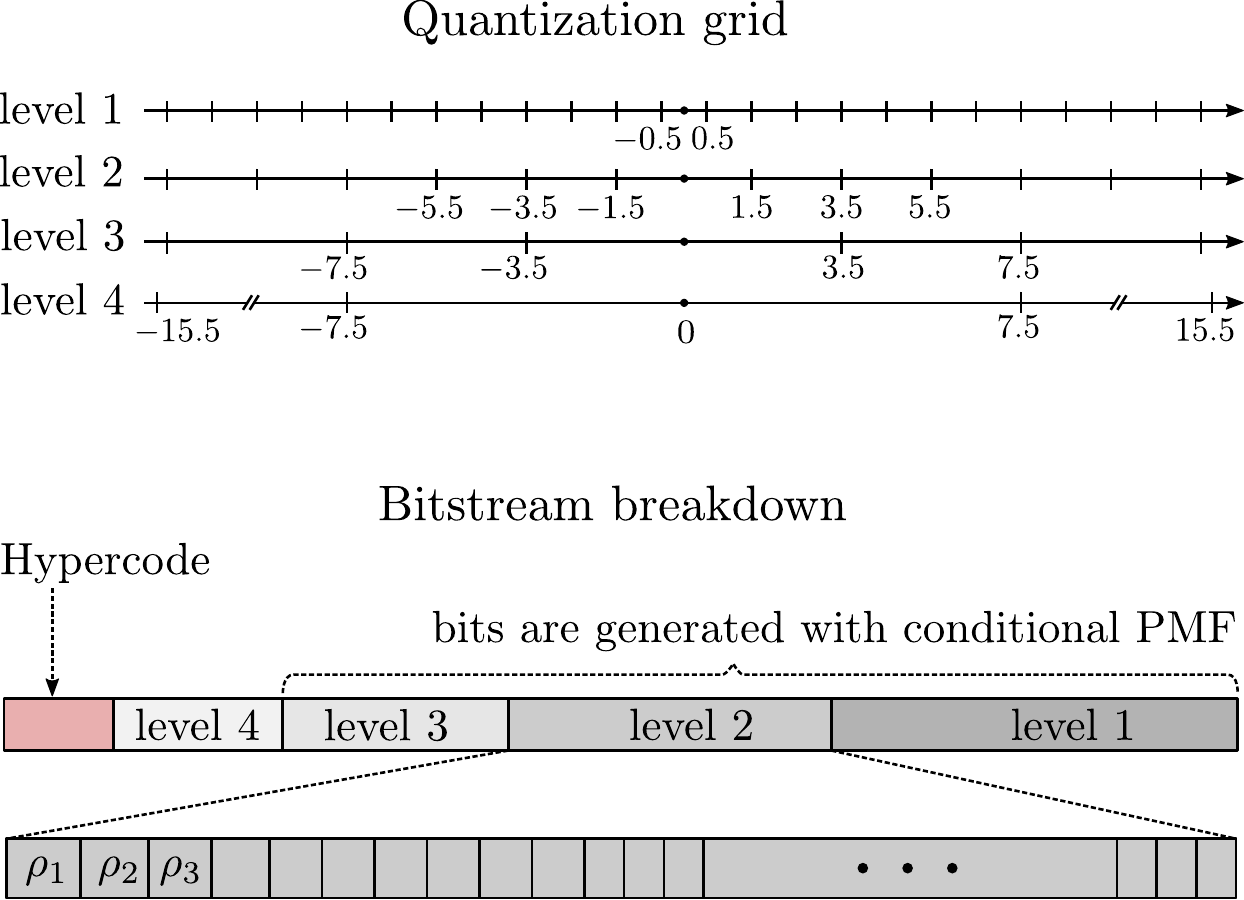}
  \caption{Quantization levels and breakdown of components in the bitstream.}
  \label{fig:pronc_illustration}
\end{subfigure}
\begin{subfigure}{.3\textwidth}
  \centering
  \includegraphics[width=\linewidth]{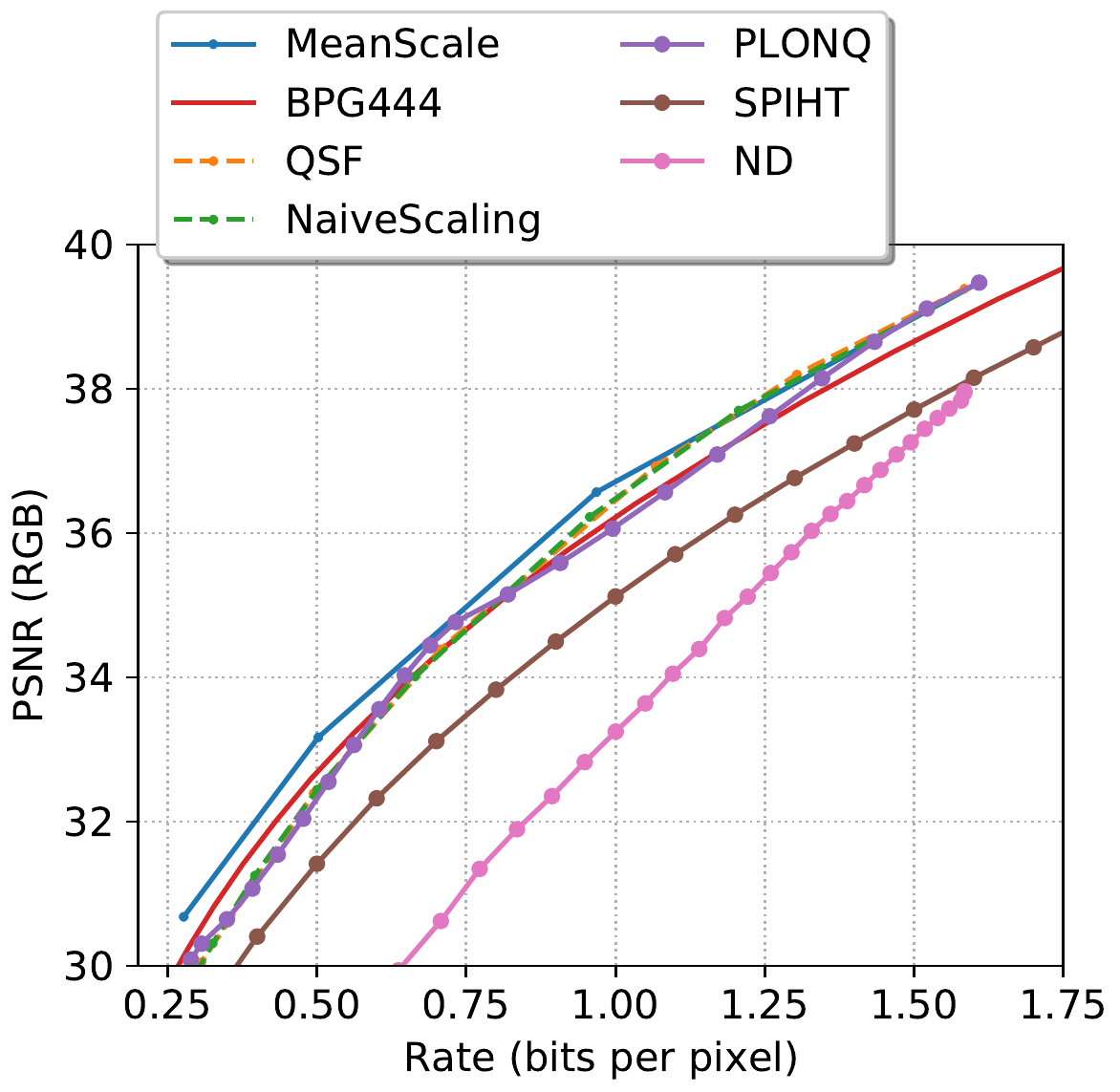}
  \caption{  Rate-distortion comparison on \\JPEG AI testset\cite{jpegai_test}}
  \label{fig:rd}
\end{subfigure}%
\begin{subfigure}{.3\textwidth}
  \centering
  \includegraphics[width=\linewidth]{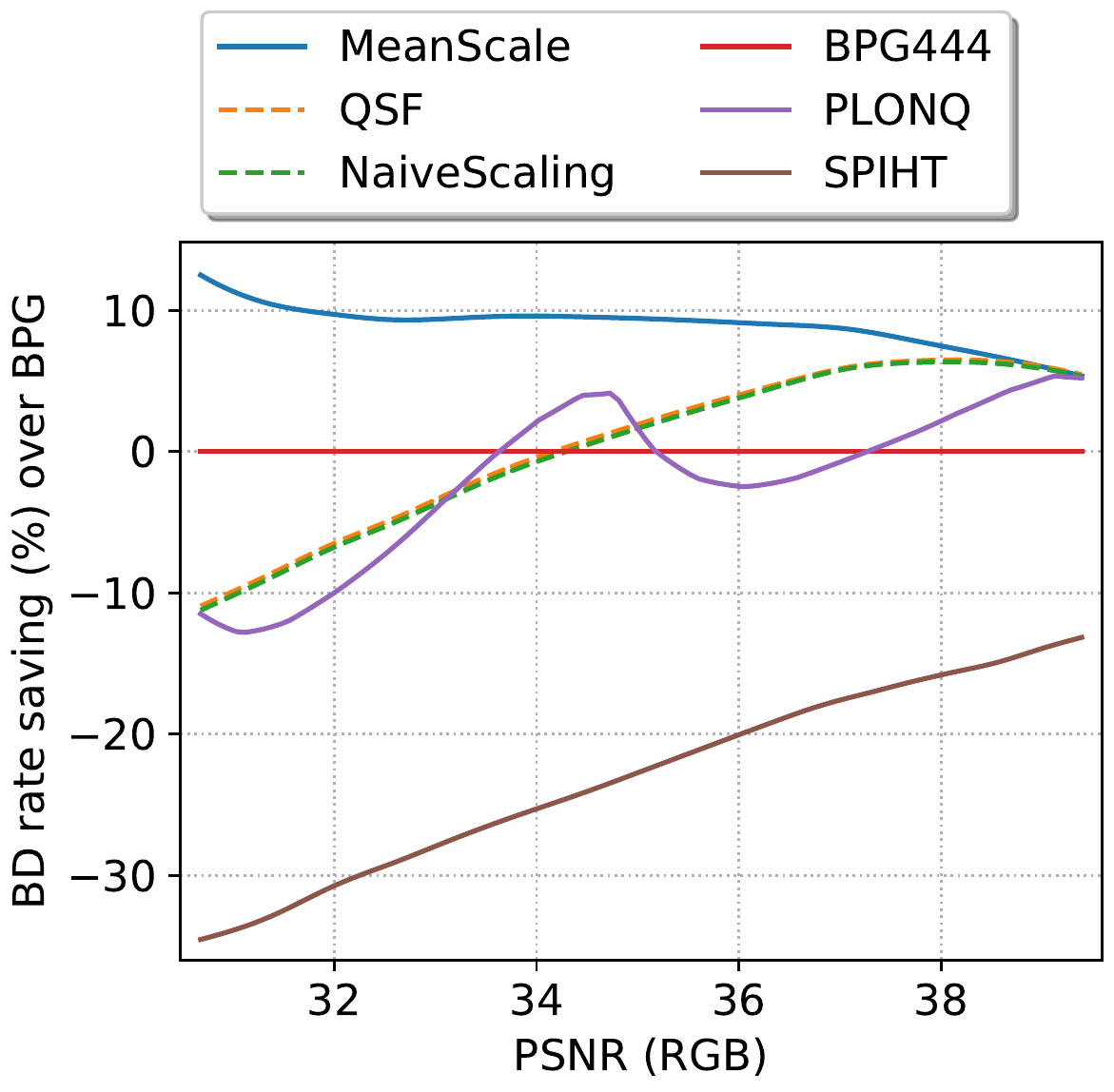}
  \caption{  BD-rate savings relative to BPG444 on \\JPEG AI testset\cite{jpegai_test}}
  \label{fig:bd}
\end{subfigure}
\caption{Illustration of PLONQ, and performance comparison of PLONQ and baseline schemes described in Table~\ref{tb:baseline}.}
\label{fig:fig}
\end{figure*}

We experimented with 6 combinations of coding unit and sorting criterion, and find the one with the best R-D performance. The latent ordering results on the JPEG AI testset~\cite{jpegai_test} is shown in Fig~\ref{fig:latent_ordering}.  It is natural to choose channels as the coding unit because each channel is usually considered as a learned feature map in ConvNet codec. Indeed channel ordering has been studied in ordered representation learning with nested dropout of latent channels~\cite{rippel2014learning} and progressive decoding  with a channel-wise auto-regressive prior model (Section $G$ in the appendix of~\cite{charm}). With channel-wise ordering, the more principled sorting criterion $\frac{\Delta D}{\Delta R}$ performs slightly better than the alternatives while being much more expensive to calculate. When using a single element as coding unit, $\frac{\Delta D}{\Delta R}$ becomes too expensive to calculate, but sorting latent elements by $\sigma$ performs much better than sorting channels by any criterion, and does not require to transmit the latent ordering, thus sorting element units by $\sigma$ is chosen for latent ordering in PLONQ.

\section{Experiment}
\label{sec:exp}
% In this section, we show implementation details of Naive Scaling, a plug-and-play variable bitrate solution and PLONQ, our proposed progressive coding scheme.
% We then evaluate their performance compared with a set of baseline solutions listed in Table~\ref{tb:baseline}.
In this section, we show experimental results of Naive Scaling, a plug-and-play variable bitrate solution and PLONQ, our proposed progressive coding scheme, compared against a set of baseline solutions listed in Table~\ref{tb:baseline}.

%and discuss their performance through experimental studies.
% and 
%conduct experiments to 

% \subsubsection*{Naive Scaling}
% Following the discussion in Section~\ref{sec:scaling}, a simple way to achieve variable bitrate with a pretrained high bitrate hyperprior model is to apply a single scaling factor $s$ across all the latents channels, for which the rate-distortion trade-off is steered by changing the value of $s$. 
% In contrast to the learnable per-channel scaling factor scheme introduced in QSF \cite{chen2020variable}, no training is involved \yl{in naive scaling}.

% Interestingly, as is shown in Fig.~\ref{fig:rd} and \ref{fig:bd}, the performance of Naive Scaling is indistinguishable to QSF, even though being much simpler, and they both outperform BPG444 at high bitrate regime. This implies simply changing the quantization bin size in the latent domain is an effective way to trade off distortion and rate in the source image domain, which leads us to the idea of exploring progressiveness across quantization levels. \yl{which motivates us to explore progressiveness across quantization levels.}

% \subsubsection*{PLONQ}
%For Native Scaling, ... 
For PLONQ, we use four quantization levels as the basis for nested quantization, and a mean-scale hyperprior (\cite{meanscale} without spatial autoregresssive context) trained with $\beta=1e-4$ as the base model. 
%To ease the computation of the conditional probability mass function in Eq~\eqref{cond_prob}, 
We enforce the quantization grid to be fully nested in the sense that a coarser quantization grid is always aligned with a finer one. 
In practice, we find it beneficial to adopt an uneven quantization grid design, where the center bin size can be different from the other bin sizes. 
%This led to a design of quantization levels with uneven quantization bins, and
Through experiments we identify the set of quantization levels illustrated in  Fig~\ref{fig:pronc_illustration} as a good design choice\footnote{these four quantization levels form a nested set of deadzone quantizers \cite{sullivan}}. 

% With nested quantization alone the granularity of progressiveness is quite limited -- there are only four bitrate options corresponds to the four quantization levels. 
To favor fine-grained rate-control, 
% it is essential to achieve incremental bitrate increase in between two quantization levels. This is enabled by computing 
PLONQ computes an image-specific ordering $\rho$ of latent elements 
%with algorithms described in Section~\ref{sec:ordering},
and then incrementally updates the latent with finer quantization level element by element following the order. 
Based on the analysis in Section~\ref{sec:ordering}, we adopt the order computed by sorting the per-element standard deviation $\sigma$ from the output of the hyper decoder. 
An illustration of breakdown of the bitstream can be found in lower half of Fig~\ref{fig:pronc_illustration}. Note that the use of image-specific latent element ordering incurs no extra bit overhead, since the ordering can be derived after hypercode is obtained.
% , encode the order as part of the bitstream,
% \footnote{In experiment we encode the order of $N$ channels with $N\lceil\log(N) \rceil$ bits.} 
% \footnote{Alternatively we could use a held-out dataset to compute a model-specific latent element order and save the overhead of transmitting the ordered sequence, but in practice we find the performance benefit of per-image channel order greatly outweighs this overhead. }

\begin{table}[h]
\small	
\caption{\label{tb:baseline}Baseline image compression schemes}
\begin{tabular}{@{}l p{6.4cm}@{}}
\toprule Label & Description\\
\midrule
\midrule
MeanScale & Mean-Scale hyperprior model (\cite{meanscale} without spatial autoregresssive context). Four models trained separately with $\beta$=1e-4, 3e-4, 1e-3, and 3e-3 for 2 million steps.\\
\midrule
BPG444 & Better Portable Graphics\cite{BPG} (HEVC All-intra) %\yz{colorspace?}
\\
QSF & Quality Scaling Factor\cite{chen2020variable}. 8 sets of learnable per-channel latent scaling  factors trained from $\beta=$1e-4 to 5e-2 on a pre-trained and freezed MeanScale($\beta$=1e-4) model.\\
NaiveScaling & Constant latent scaling factors $s=1,2,\ldots,8$ applied to a pre-trained MeanScale($\beta$=1e-4) model.\\
\midrule
ND & MeanScale($\beta$=1e-4) with Nested Dropout\cite{rippel2014learning} training.\\
SPIHT & Set Partitioning In Hierarchical Trees\cite{spiht}\\
\bottomrule
\end{tabular}
\end{table}

\vspace{-0.2in}

\subsection{Performance evaluation}
We evaluate the performance of PLONQ against an array of baseline schemes outlined in Table~\ref{tb:baseline}, which divides into three categories (as delimited in the table): (1) multi-model compression scheme (2) non-progressive single model variable bitrate scheme (3) progressive compression scheme. Among them, BPG444 and SPIHT are traditional codecs and the rest are learning-based codecs.

Fig~\ref{fig:rd} and \ref{fig:bd} show the RD curve (bpp vs PSNR) as well as the BD-rate\cite{BD} saving relative to BPG444 computed on JPEG AI dataset \cite{jpegai_test}. The markers on the progressive solution RD curves indicate the achievable rate options. As shown in the two figures, PLONQ is significantly better than the learned nested dropout \cite{rippel2014learning}. Further PLONQ outperforms the well-known conventional progressive coding scheme SPIHT \cite{spiht} by a considerable margin uniformly across the whole bpp range, and it is even competitive to the non-progressive coding scheme BPG444 at high bpp regime. The same observation is 
%further validated
made
on Kodak and Tecnick dataset, for which the results can be found in the appendix. 
%To furthre
% Alternatively we could use a held-out dataset to compute a model-specific channel order and saves the overhead of transmitting the ordered sequence, but in practise we find the performance benefit of per-image channel order greatly outweigh this overhead. 

%In this section, we first show that  we perform experimental studies to demonstrate the effectiveness of the combined nested quantization and channel ordering based progressive coding. Baselines schemes are summarized in Table~\ref{tb:baseline} and can be divided into three categories 

%Four quantization levels with nested grid are used in the experiments. For each quantization level, the encoded bits for different latent channels are arranged according to channel order transmitted in earlier portion of the bitstream.

%We first train a mean-scale hyper-prior model (\cite{meanscale} without autoregressive component) 

%\yy{Remake the figure for jpegai and start writing} \cite{jpegai_test}

% \begin{figure}[htb]
%   \centering
%   \includegraphics[width=0.7\linewidth]{figures/jpegai_test_BD.pdf}
%   \caption{JPEGAI test BD}
%   \label{fig:tecnick_test_BD}
% \end{figure}

% \begin{figure}[htb]
%   \centering
%   \includegraphics[width=0.7\linewidth]{figures/jpegai_test_RD.pdf}
%   \caption{JPEGAI test RD}
%   \label{fig:tecnick_test_BD}
% \end{figure}

% \note{0.75 page for this section}
\section{Conclusion}
\label{sec:conclusion}
We have introduced PLONQ, a learning based progressive image codec using nested quantization and latent ordering. It uniformly outperforms the existing wavelet-based progressive image codec SPIHT and matches or even outperforms BPG444 in the high bitrate region. The effectiveness of PLONQ proves itself to be a solid baseline for learning based progressive codec and gives the first glimpse of its potential to enable progressive neural video coding.

% References should be produced using the bibtex program from suitable
% BiBTeX files (here: strings, refs, manuals). The IEEEbib.bst bibliography
% style file from IEEE produces unsorted bibliography list.
% -------------------------------------------------------------------------

\clearpage
\bibliographystyle{IEEEbib}
\bibliography{refs}

\newpage
\appendix
\section{Dataset preprocessing}
The following two image in the original JPEG AI test dataset incur more GPU memory than is available to us for mean-scale hyper-prior inference, and thus we cropped it to \texttt{3680x2160} (cropped from top) for all our experiments.

     \texttt{00014\_TE\_3680x2456.png}
     
     \texttt{00015\_TE\_3680x2456.png}
     
\section{Evaluation of BPG444 and SPIHT}

BPG444 \cite{BPG} results are generated with the following command:

\texttt{bpgenc -e x265 -q [qp] -f 444}

\texttt{-o [output file] [input file]}

The results on Kodak dataset (shown in \ref{fig:kodak_rd}) is validated against that provided in\\ \texttt{https://github.com/tensorflow/compression/}\\ \texttt{blob/master/results/image\_compression/}\\ \texttt{kodak/PSNR\_sRGB\_RGB/bpg444.txt}

SPIHT\cite{spiht} results are obtained by encoding each image with a target bpp of 2.0 and decoding by truncating the bitstream by a step size of 0.1bpp. The curves are generated by averaging PSNR across images for each of the fixed bpps. 

\section{Additional experiment results on Kodak and Tecnick testset}

\begin{figure}[h]
  \centering
  \includegraphics[width=0.9\linewidth]{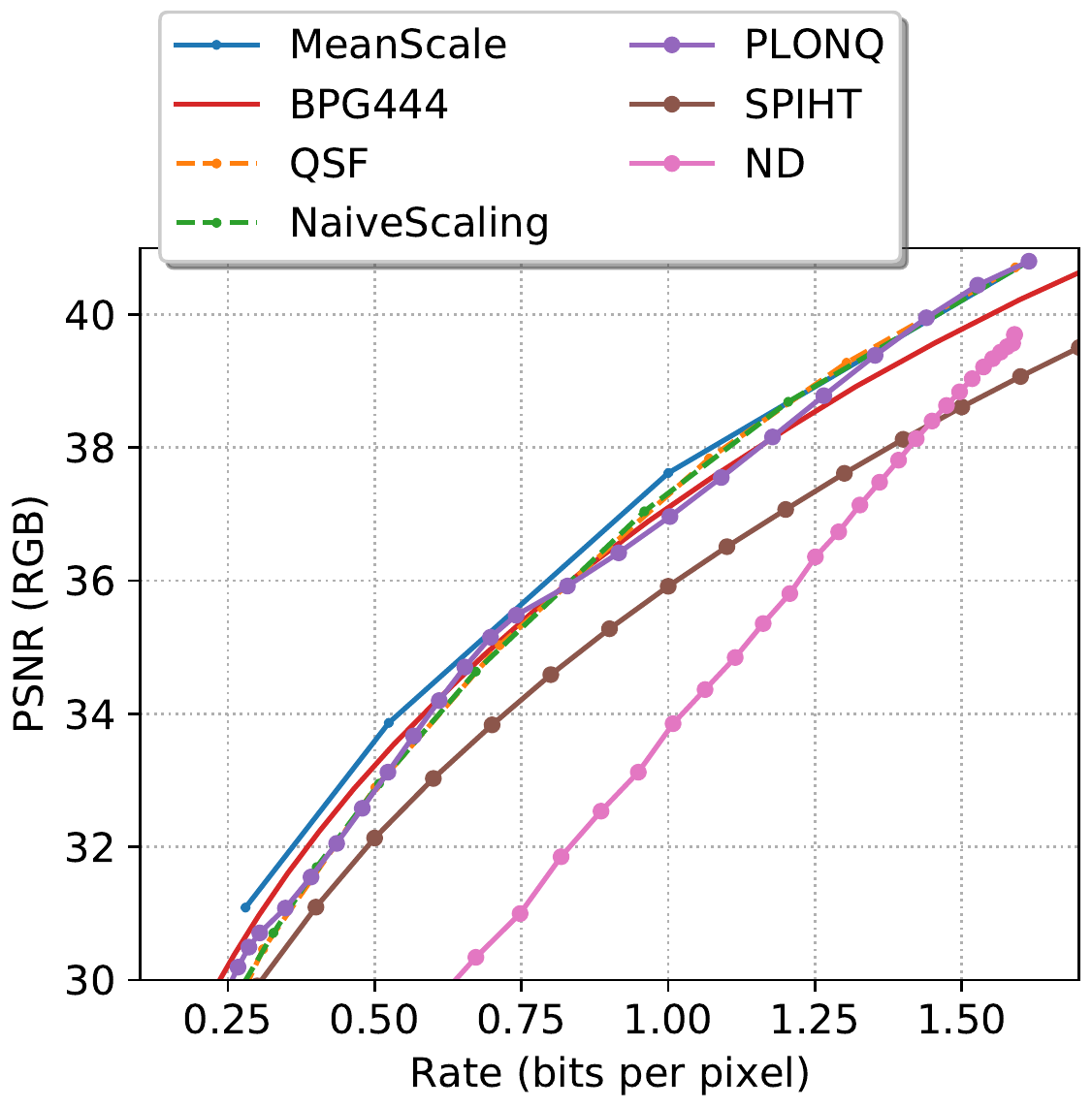}
  \caption{Rate-distortion comparison on Kodak.}
  \label{fig:kodak_rd}
\end{figure}

\begin{figure}[h]
  \centering
  \includegraphics[width=0.9\linewidth]{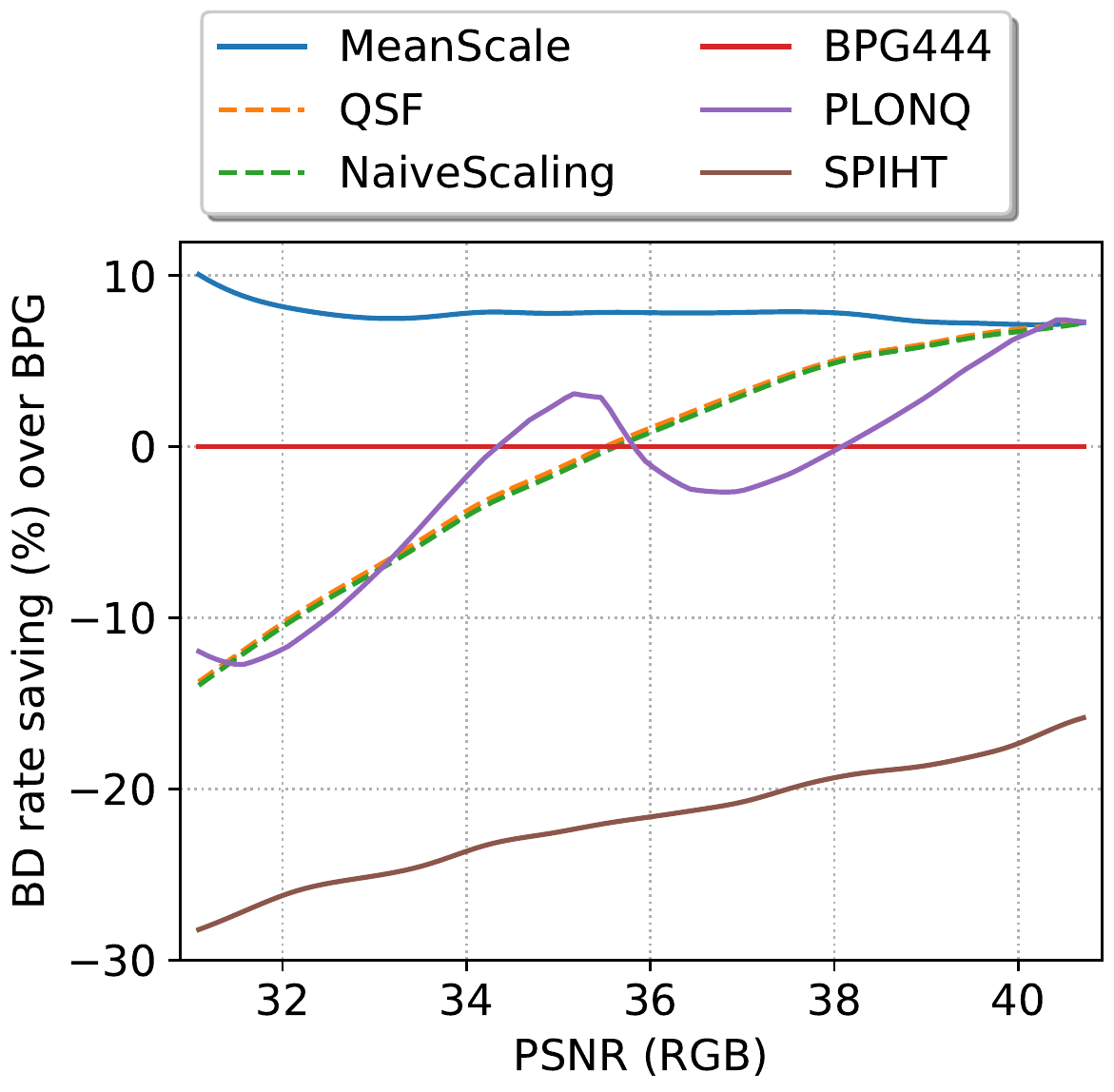}
  \caption{BD-rate savings relative to BPG444 on Kodak.}
  \label{fig:kodak_bd}
\end{figure}

\begin{figure}[h]
  \centering
  \includegraphics[width=0.9\linewidth]{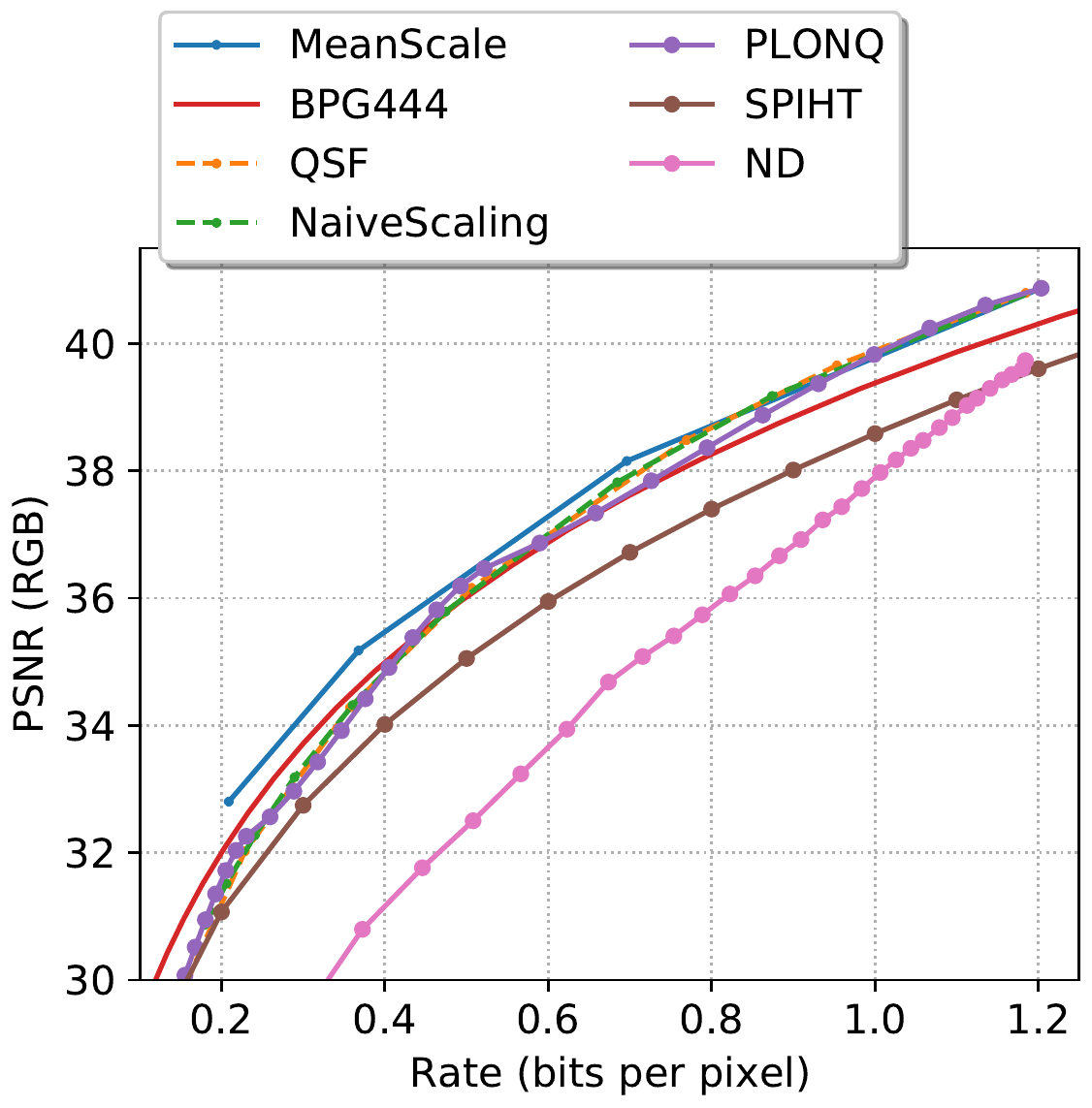}
  \caption{Rate-distortion comparison on Tecnick testset.}
  \label{fig:tecnick_rd}
\end{figure}

\begin{figure}[h]
  \centering
  \includegraphics[width=0.9\linewidth]{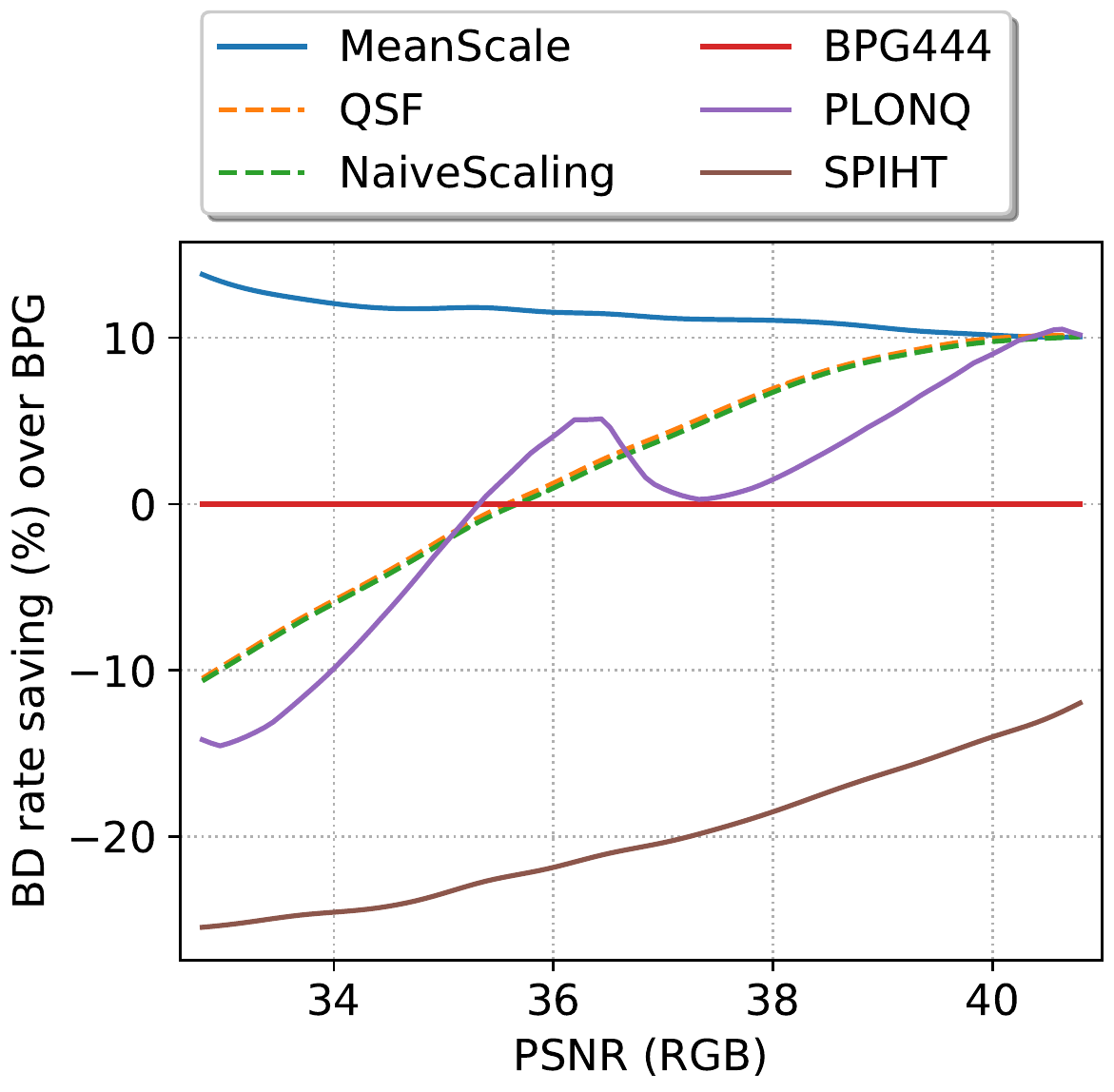}
  \caption{BD-rate savings relative to BPG444 on Tecnick testset.}
  \label{fig:tecnick_bd}
\end{figure}

Fig~\ref{fig:kodak_rd} and \ref{fig:kodak_bd} show the performance comparison of PLONQ and baseline schemes described in Table~\ref{tb:baseline} on Kodak; Fig~\ref{fig:tecnick_rd} and \ref{fig:tecnick_bd} shows the results on Tecnick test dataset (100 images with 1200 x 1200 resolution).

% \begin{figure}[h]
% \begin{subfigure}{.5\textwidth}
%   \centering
%   \includegraphics[width=0.7\linewidth]{figures/kodak_RD.pdf}
%   \caption{ \centering Rate-distortion comparison on Kodak}
% \end{subfigure}\\
% \begin{subfigure}{.5\textwidth}
%   \centering
%   \includegraphics[width=0.7\linewidth]{figures/kodak_BD.pdf}
%   \caption{ \centering BD-rate savings relative to BPG444 on Kodak}
% \end{subfigure}
% \caption{Performance comparison of PLONQ and baseline schemes described in Table~\ref{tb:baseline} on Kodak.}\label{fig:kodak}
% \end{figure}

% \begin{figure}[h]
% \begin{subfigure}{.5\textwidth}
%   \centering
%   \includegraphics[width=0.7\linewidth]{figures/tecnick_test_RD.pdf}
%   \caption{ \centering Rate-distortion comparison on Tecnick testset}
% \end{subfigure}\\
% \begin{subfigure}{.5\textwidth}
%   \centering
%   \includegraphics[width=0.7\linewidth]{figures/tecnick_test_BD.pdf}
%   \caption{ \centering BD-rate savings relative to BPG444 on Tecnick testset}
% \end{subfigure}
% \caption{Performance comparison of PLONQ and baseline schemes described in Table~\ref{tb:baseline} on Tecnick testset (100 images with 1200 x 1200 resolution).}\label{fig:tecnick}
% \end{figure}

\end{document}